\documentclass[11pt,a4paper]{article}
\usepackage[hyperref]{acl2017}
\usepackage{times}
\usepackage{latexsym}
\usepackage{graphicx}
\usepackage{url}

\aclfinalcopy 

\title{A Biomedical Information Extraction Primer for NLP Researchers}

\author{Surag Nair \\
  Indian Institute of Technology Delhi \\
  {\tt ee1130504@iitd.ac.in} \\}

\date{}

\begin{document}
\maketitle
\begin{abstract}
Biomedical Information Extraction is an exciting field at the crossroads of Natural Language Processing, Biology and Medicine. It encompasses a variety of different tasks that require application of state-of-the-art NLP techniques, such as NER and Relation Extraction. This paper provides an overview of the problems in the field and discusses some of the techniques used for solving them.
\end{abstract}

\section{Introduction}
The explosion of available scientific articles in the Biomedical domain has led to the rise of Biomedical Information Extraction (BioIE). BioIE systems aim to extract information from a wide spectrum of articles including medical literature, biological literature, electronic health records, etc. that can be used by clinicians and researchers in the field. Often the outputs of BioIE systems are used to assist in the creation of databases, or to suggest new paths for research. For example, a ranked list of interacting proteins that are extracted from biomedical literature, but are not present in existing databases, can allow researchers to make informed decisions about which protein/gene to study further. Interactions between drugs are necessary for clinicians who simultaneously administer multiple drugs to their patients. A database of diseases, treatments and tests is beneficial for doctors consulting in complicated medical cases.

The main problems in BioIE are similar to those in Information Extraction:
\begin{enumerate}
    \item Named Entity Recognition
    \item Relation Extraction
    \item Event Extraction
\end{enumerate}

This paper discusses, in each section, various methods that have been adopted to solve the listed problems. Each section also highlights the difficulty of Information Extraction tasks in the biomedical domain.

This paper is intended as a primer to Biomedical Information Extraction for current NLP researchers. It aims to highlight the diversity of the various techniques from Information Extraction that have been applied in the Biomedical domain. The state of biomedical text mining is reviewed regularly. For more extensive surveys, consult \cite{liu2016learning}, \cite{aggarwal2012mining}, \cite{zweigenbaum2007frontiers}.

\section{Named Entity Recognition and Fact Extraction}
Named Entity Recognition (NER) in the Biomedical domain usually includes recognition of entities such as proteins, genes, diseases, treatments, drugs, etc. Fact extraction involves extraction of Named Entities from a corpus, usually given a certain ontology. When compared to NER in the domain of general text, the biomedical domain has some characteristic challenges:
\begin{enumerate}
    \item Synonymy: the same biomedical entity is often known by different names. E.g.  ``cyclin-dependent kinase inhibitor p27'' and ``‘p27kip1'' are the same proteins, ``heart attack'' and ``myocardial infarcation'' refer to the same medical problem.
    \item Abbreviations: The literature is rich with ambiguous abbreviations: ``RA'' can refer to ``right atrium'', ``rheumatoid arthritis'', ``renal artery'' or several other concepts \cite{pakhomov2002semi}
    \item Entity names are subject to many variants, and also change over time
\end{enumerate}

Some of the earliest systems were heavily dependent on hand-crafted features. The method proposed in \cite{fukudatoward} for recognition of protein names in text does not require any prepared dictionary. The work gives examples of diversity in protein names and lists multiple rules depending on simple word features as well as POS tags.

\cite{de2011machine} adopt a machine learning approach for NER. Their NER system extracts medical problems, tests and treatments from discharge summaries and progress notes. They use a semi-Conditional Random Field (semi-CRF) \cite{sarawagi2004semi} to output labels over all tokens in the sentence. They use a variety of token, context and sentence level features. They also use some concept mapping features using existing annotation tools, as well as Brown clustering to form 128 clusters over the unlabelled data. The dataset used is the i2b2 2010 challenge dataset. Their system achieves an F-Score of 0.85. \cite{tang2014evaluating} is an incremental paper on NER taggers. It uses 3 types of word-representation techniques (Brown clustering, distributional clustering, word vectors) to improve performance of the NER Conditional Random Field tagger, and achieves marginal F-Score improvements.

\cite{movshovitz2012bootstrapping} propose a boostrapping mechanism to bootstrap biomedical ontologies using NELL \cite{carlson2010toward}, which uses a coupled semi-supervised bootstrapping approach to extract facts from text, given an ontology and a small number of ``seed'' examples for each category. This interesting approach (called BioNELL) uses an ontology of over 100 categories. In contrast to NELL, BioNELL does not contain any relations in the ontology. BioNELL is motivated by the fact that a lot of scientific literature available online is highly reliable due to peer-review. The authors note that the algorithm used by NELL to bootstrap fails in BioNELL due to ambiguities in biomedical literature, and heavy semantic drift. One of the causes for this is that often common words such as ``white'', ``dad'', ``arm'' are used as names of genes- this can easily result in semantic drift in one iteration of the bootstrapping. In order to mitigate this, they use Pointwise Mutual Information scores for corpus level statistics, which attributes a small score to common words. In addition, in contrast to NELL, BioNELL only uses high instances as seeds in the next iteration, but adds low ranking instances to the knowledge base. Since evaluation is not possible using Mechanical Turk or a small number of experts (due to the complexity of the task), they use Freebase \cite{bollacker2008freebase}, a knowledge base that has some biomedical concepts as well. The lexicon learned using BioNELL is used to train an NER system. The system shows a very high precision, thereby showing that BioNELL learns very few ambiguous terms.

More recently, deep learning techniques have been developed to further enhance the performance of NER systems. \cite{li2015exploring} explore recurrent neural networks for the problem of NER in biomedical text.

\section{Relation Extraction}
In Biomedical Information Extraction, Relation Extraction involves finding related entities of many different kinds. Some of these include protein-protein interactions, disease-gene relations and drug-drug interactions. Due to the explosion of available biomedical literature, it is impossible for one person to extract relevant relations from published material. Automatic extraction of relations assists in the process of database creation, by suggesting potentially related entities with links to the source article. For example, a database of drug-drug interactions is important for clinicians who administer multiple drugs simultaneously to their patients- it is imperative to know if one drug will have an adverse effect on the other. A variety of methods have been developed for relation extractions, and are often inspired by Relation Extraction in NLP tasks. These include rule-based approaches, hand-crafted patterns, feature-based and kernel machine learning methods, and more recently deep learning architectures. Relation Extraction systems over Biomedical Corpora are often affected by noisy extraction of entities, due to ambiguities in names of proteins, genes, drugs etc.

\cite{blaschke2001can} was one of the first large scale Information Extraction efforts to study the feasibility of extraction of protein-protein interactions (such as ``protein A activates protein B") from Biomedical text. Using 8 hand-crafted regular expressions over a fixed vocabulary, the authors were able to achieve a recall of 30\% for interactions present in The Dictionary of Interacting Proteins (DIP) from abstracts in Medline. The method did not differentiate between the type of relation. The reasons for the low recall were the inconsistency in protein nomenclature, information not present in the abstract, and due to specificity of the hand-crafted patterns. On a small subset of extracted relations, they found that about 60\% were true interactions between proteins not present in DIP. 

\cite{bunescu2006integrating} combine sentence level relation extraction for protein interactions with corpus level statistics. Similar to \cite{blaschke2001can}, they do not consider the type of interaction between proteins- only whether they interact in the general sense of the word. They also do not differentiate between genes and their protein products (which may share the same name). They use Pointwise Mutual Information (PMI) for corpus level statistics to determine whether a pair of proteins occur together by chance or because they interact. They combine this with a confidence aggregator that takes the maximum of the confidence of the extractor over all extractions for the same protein-pair. The extraction uses a subsequence kernel based on \cite{bunescu2005subsequence}. The integrated model, that combines PMI with aggregate confidence, gives the best performance. Kernel methods have widely been studied for Relation Extraction in Biomedical Literature. Common kernels used usually exploit linguistic information by utilising kernels based on the dependency tree \cite{liu2013new}, \cite{zhang2012hash}, \cite{patra2013kernel}.

\cite{chun2006extraction} look at the extraction of diseases and their relevant genes. They use a dictionary from six public databases to annotate genes and diseases in Medline abstracts. In their work, the authors note that when both genes and diseases are correctly identified, they are related in 94\% of the cases. The problem then reduces to filtering incorrect matches using the dictionary, which occurs due to false positives resulting from ambiguities in the names as well as ambiguities in abbreviations. To this end, they train a Max-Ent based NER classifier for the task, and get a 26\% gain in precision over the unfiltered baseline, with a slight hit in recall. They use POS tags, expanded forms of abbreviations, indicators for Greek letters as well as suffixes and prefixes commonly used in biomedical terms.

\cite{bui2014novel} adopt a supervised feature-based approach for the extraction of drug-drug interaction (DDI) for the DDI-2013 dataset \cite{herrero2013ddi}. They partition the data in subsets depending on the syntactic features, and train a different model for each. They use lexical, syntactic and verb based features on top of shallow parse features, in addition to a hand-crafted list of trigger words to define their features. An SVM classifier is then trained on the feature vectors, with a positive label if the drug pair interacts, and negative otherwise. Their method beats other systems on the DDI-2013 dataset. Some other feature-based approaches are described in \cite{leaman2015tmchem}, \cite{bui2011hybrid}.

Distant supervision methods have also been applied to relation extraction over biomedical corpora. In \cite{liu2014relation}, 10,000 neuroscience articles are distantly supervised using information from UMLS Semantic Network to classify brain-gene relations into geneExpression and otherRelation. They use lexical (bag of words, contextual) features as well as syntactic (dependency parse features). They make the ``at-least one'' assumption, i.e. at least one of the sentences extracted for a given entity-pair contains the relation in database. They model it as a multi-instance learning problem and adopt a graphical model similar to \cite{hoffmann2011knowledge}. They test using manually annotated examples. They note that the F-score achieved are much lesser than that achieved in the general domain in \cite{hoffmann2011knowledge}, and attribute to generally poorer performance of NER tools in the biomedical domain, as well as less training examples. \cite{thomas2011learning} explore distant supervision methods for protein-protein interaction extraction.

More recently, deep learning methods have been applied to relation extraction in the biomedical domain. One of the main advantages of such methods over traditional feature or kernel based learning methods is that they require minimal feature engineering. In \cite{jiang2016general}, skip-gram vectors \cite{mikolov2013distributed} are trained over 5.6Gb of unlabelled text. They use these vectors to extract protein-protein interactions by converting them into features for entities, context and the entire sentence. Using an SVM for classification, their method is able to outperform many kernel and feature based methods over a variety of datasets.

\cite{sahu2016relation} follow a similar method by using word vectors trained on PubMed articles. They use it for the task of relation extraction from clinical text for entities that include problem, treatment and medical test. For a given sentence, given labelled entities, they predict the type of relation exhibited (or None) by the entity pair. These types include ``treatment caused medical problem'', ``test conducted to investigate medical problem'', ``medical problem indicates medical problems'', etc. They use a Convolutional Neural Network (CNN) followed by feedforward neural network architecture for prediction. In addition to pre-trained word vectors as features, for each token they also add features for POS tags, distance from both the entities in the sentence, as well BIO tags for the entities. Their model performs better than a feature based SVM baseline that they train themselves.

The BioNLP'16 Shared Tasks has also introduced some Relation Extraction tasks, in particular the BB3-event subtask that involves predicting whether a ``lives-in'' relation holds for a Bacteria in a location. Some of the top performing models for this task are deep learning models. \cite{mehryary2016deep} train word embeddings with six billions words of scientific texts from PubMed. They then consider the shortest dependency path between the two entities (Bacteria and location). For each token in the path, they use word embedding features, POS type embeddings and dependency type embeddings. They train a unidirectional LSTM \cite{hochreiter1997long} over the dependency path, that achieves an F-Score of 52.1\% on the test set.

\cite{li2016biomedical} improve the performance by making modifications to the above model. Instead of using the shortest dependency path, they modify the parse tree based on some pruning strategies. They also add feature embeddings for each token to represent the distance from the entities in the shortest path. They then train a Bidirectional LSTM on the path, and obtain an F-Score of 57.1\%.

The recent success of deep learning models in Biomedical Relation Extraction that require minimal feature engineering is promising. This also suggests new avenues of research in the field. An approach as in \cite{zeng2015distant} can be used to combine multi-instance learning and distant supervision with a neural architecture.

\section{Event Extraction}
Event Extraction in the Biomedical domain is a task that has gained more importance recently. Event Extraction goes beyond Relation Extraction. In Biomedical Event Extraction, events generally refer to a change in the state of biological molecules such as proteins and DNA. Generally, it includes detection of targeted event types such as gene expression, regulation, localisation and transcription. Each event type in addition can have multiple arguments that need to be detected. An additional layer of complexity comes from the fact that events can also be arguments of other events, giving rise to a nested structure. This helps to capture the underlying biology better \cite{aggarwal2012mining}. Detecting the event type often involves recognising and classifying trigger words. Often, these words are verbs such as ``activates'', ``inhibits'', ``phosphorylation'' that may indicate a single, or sometimes multiple event types. In this section, we will discuss some of the successful models for Event Extraction in some detail.

\begin{figure}[h!]
  \includegraphics[width=\linewidth]{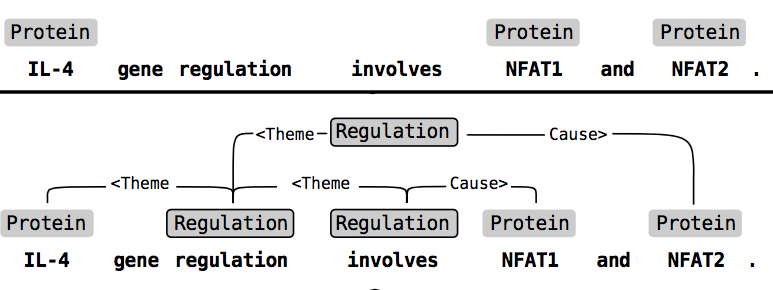}
  \caption{An example of an input sentence with annotations, and expected output of the event extraction system (Borrowed from \cite{bjorne2009extracting})}
\end{figure}

Event Extraction gained a lot of interest with the availability of an annotated corpus with the BioNLP'09 Shared Task on Event Extraction \cite{kim2008corpus}. The task involves prediction of trigger words over nine event types such as expression, transcription, catabolism, binding, etc. given only annotation of named entities (proteins, genes, etc.). For each event, its class, trigger expression and arguments need to be extracted. Since the events can be arguments to other events, the final output in general is a graph representation with events and named entities as nodes, and edges that correspond to event arguments. \cite{bjorne2009extracting} present a pipeline based method that is heavily dependent on dependency parsing. Their pipeline approach consists of three steps: trigger detection, argument detection and semantic post-processing. While the first two components are learning based systems, the last component is a rule based system. For the BioNLP'09 corpus, only 5\% of the events span multiple sentences. Hence the approach does not get affected severely by considering only single sentences. It is important to note that trigger words cannot simply be reduced to a dictionary lookup. This is because a specific word may belong to multiple classes, or may not always be a trigger word for an event. For example, ``activate'' is found to not be a trigger word in over 70\% of the cases. A multi-class SVM is trained for trigger detection on each token, using a large feature set consisting of semantic and syntactic features. It is interesting to note that the hyperparameters of this classifier are optimised based on the performance of the entire end-to-end system.

For the second component to detect arguments, labels for edges between entities must be predicted. For the BioNLP'09 Shared Task, each directed edge from one event node to another event node, or from an event node to a named entity node are classified as ``theme'', ``cause'', or None. The second component of the pipeline makes these predictions independently. This is also trained using a multi-class SVM which involves heavy use of syntactic features, including the shortest dependency path between the nodes. The authors note that the precision-recall choice of the first component affects the performance of the second component: since the second component is only trained on Gold examples, any error by the first component will lead to a cascading of errors. The final component, which is a semantic post-processing step, consists of rules and heuristics to correct the output of the second component. Since the edge predictions are made independently, it is possible that some event nodes do not have any edges, or have an improper combination of edges. The rule based component corrects these and applies rules to break directed cycles in the graph, and some specific heuristics for different types of events. The final model gives a cumulative F-Score of 52\% on the test set, and was the best model on the task.

\cite{poon2010joint} note that previous approaches on the task suffer due to the pipeline nature and the propagation of errors. To counter this, they adopt a joint inference method based on Markov Logic Networks \cite{richardson2006markov} for the same task on BioNLP'09. The Markov Logic Network jointly predicts whether each token is a trigger word, and if yes, the class it belongs to; for each dependency edge, whether it is an argument path leading to a ``theme'' or a ``cause''. By formulating the Event Extraction problem using an MLN, the approach becomes computationally feasible and only linear in the length of the sentence. They incorporate hard constraints to encode rules such as ``an argument path must have an event'', ``a cause path must start with a regulation event'', etc. In addition, they also include some domain specific soft constraints as well as some linguistically-motivated context-specific soft constraints. In order to train the MLN, stochastic gradient descent was used. Certain heuristic methods are implemented in order to deal with errors due to syntactic parsing, especially ambiguities in PP-attachment and coordination. Their final system is competitive and comes very close to the system by \cite{bjorne2009extracting} with an average F-Score of 50\%. To further improve the system, they suggest leveraging additional joint-inference opportunities and integrating the syntactic parser better. Some other more recent models for Biomedical Event Extraction include \cite{riedel2011fast}, \cite{mcclosky2012combining}.




\section{Conclusion}
We have discussed some of the major problems and challenges in BioIE, and seen some of the diverse approaches adopted to solve them. Some interesting problems such as Pathway Extraction for Biological Systems \cite{ananiadou2010event}, \cite{rzhetsky2004geneways} have not been discussed. 

Biomedical Information Extraction is a challenging and exciting field for NLP researchers that demands application of state-of-the-art methods. Traditionally, there has been a dependence on hand-crafted features or heavily feature-engineered methods. However, with the advent of deep learning methods, a lot of BioIE tasks are seeing an improvement by adopting deep learning models such as Convolutional Neural Networks and LSTMs, which require minimal feature engineering. Rapid progress in developing better systems for BioIE will be extremely helpful for clinicians and researchers in the Biomedical domain.

\bibliography{acl2017}
\bibliographystyle{acl_natbib}

\appendix

\end{document}